\documentclass[review]{elsarticle}

\usepackage{lineno,hyperref}

\usepackage{amssymb}
\usepackage{amsmath}
\usepackage{graphicx}
\usepackage{float} 
\usepackage{subfigure}

\usepackage{booktabs}
\usepackage{multirow}
\usepackage{threeparttable}
\usepackage{color}
\usepackage{bm}
\usepackage{longtable}
\newcommand{\tabincell}[2]{\begin{tabular}{@{}#1@{}}#2\end{tabular}}

\modulolinenumbers[5]

\journal{Journal of Pattern Recognition}

\begin{document}

\begin{frontmatter}

\title{Tell-the-difference: Fine-grained Visual Descriptor via a Discriminating Referee}







\author[mymainaddress]{Shuangjie Xu\fnref{myfootnote}}
\ead{shuangjiexu@hust.edu.cn}

\author[mymainaddress]{Feng Xu\fnref{myfootnote}}
\ead{xufeng2015@hust.edu.cn}

\fntext[myfootnote]{Shuangjie Xu and Feng Xu have equal contribution to the paper.}

\author[mysecondaryaddress]{Yu Cheng}
\ead{yu.cheng@microsoft.com}

\author[mymainaddress]{Pan Zhou\corref{mycorrespondingauthor}}

\cortext[mycorrespondingauthor]{Corresponding author}
\ead{panzhou@hust.edu.cn}

\address[mymainaddress]{Huazhong University of Science and Technology, Luoyu Road 1037, Wuhan, China}
\address[mysecondaryaddress]{Microsoft Research AI, Microsoft Building 99, 14820 NE 36th St., Redmond, Wash., USA}

\begin{abstract}
In this paper, we investigate a novel problem of telling the difference between image pairs in natural language. Compared to previous approaches for single image captioning, it is challenging to fetch linguistic representation from two independent visual information. To this end, we have proposed an effective encoder-decoder caption framework based on Hyper Convolution Net. In addition, a series of novel feature fusing techniques for pairwise visual information fusing are introduced and a discriminating referee is proposed to evaluate the pipeline. Because of the lack of appropriate datasets to support this task, we have collected and annotated a large new dataset with Amazon Mechanical Turk (AMT) for generating captions in a pairwise manner (with 14764 images and 26710 image pairs in total). The dataset is the first one on the relative difference caption task that provides descriptions in free language. We evaluate the effectiveness of our model on two datasets in the field and it outperforms the state-of-the-art approach by a large margin.
\end{abstract}

\begin{keyword}
Relative difference caption, Fine-grained difference description, Discriminating Referee
\end{keyword}

\end{frontmatter}

\section{Introduction}
\label{sec:introduction}
Visual difference description has long been relatively less studied but crucial task in the visual-linguistic field, which is demanding because of the necessity for understanding the pairwise visual information. Most previous works could only detect visual differences within certain attribute domain designed by experts ~\cite{andreas2016reasoning}, which precludes its extension to new domains. 

Su et al.~\cite{su2017reasoning} focused on the phrase generation of object attributes for visual differences description between instances, utilizing data annotated by non-expert workers.
However, the discrepancies sentences generation is still challenging. What we aim for in this paper is a difference captioner that not only adapted to varied expressions from non-expert workers but also is capable of generating syntactical-flexible sentences.

Traditional caption tasks~\cite{vinyals2015show,xu2015show,you2016image} concentrated on single image caption tasks ($\bm{I} \to \bm{S}$, which is widely used in automatic photo captioning, human-computer interaction and so on). Recently, the discriminating caption task~\cite{vedantam2017context,Luo_2018_CVPR}, which utilizes image pair in training procedure to improve the quality of single image caption task, extends the caption task to the pairwise caption task $(\bm{I_1},\bm{I_2}) \to \bm{S}$. However, the discriminating caption task only has a fuzzy perception of the image difference, leading to a limited ability to tell the difference in detail. In this work, we aim to distinguish image pair with detailed attributes. For example, the discriminating captioner tells "An ultralight plane flying through a blue sky", but our model tells "Has no landing gear can be seen and is an ultralight plane", where the detailed difference receives more attention by the model than the entire scene. 

\begin{figure}[!tp]
\centering
\label{Fig.data}
\subfigure{
\label{Fig.shoes}
\includegraphics[width=0.5\textwidth]{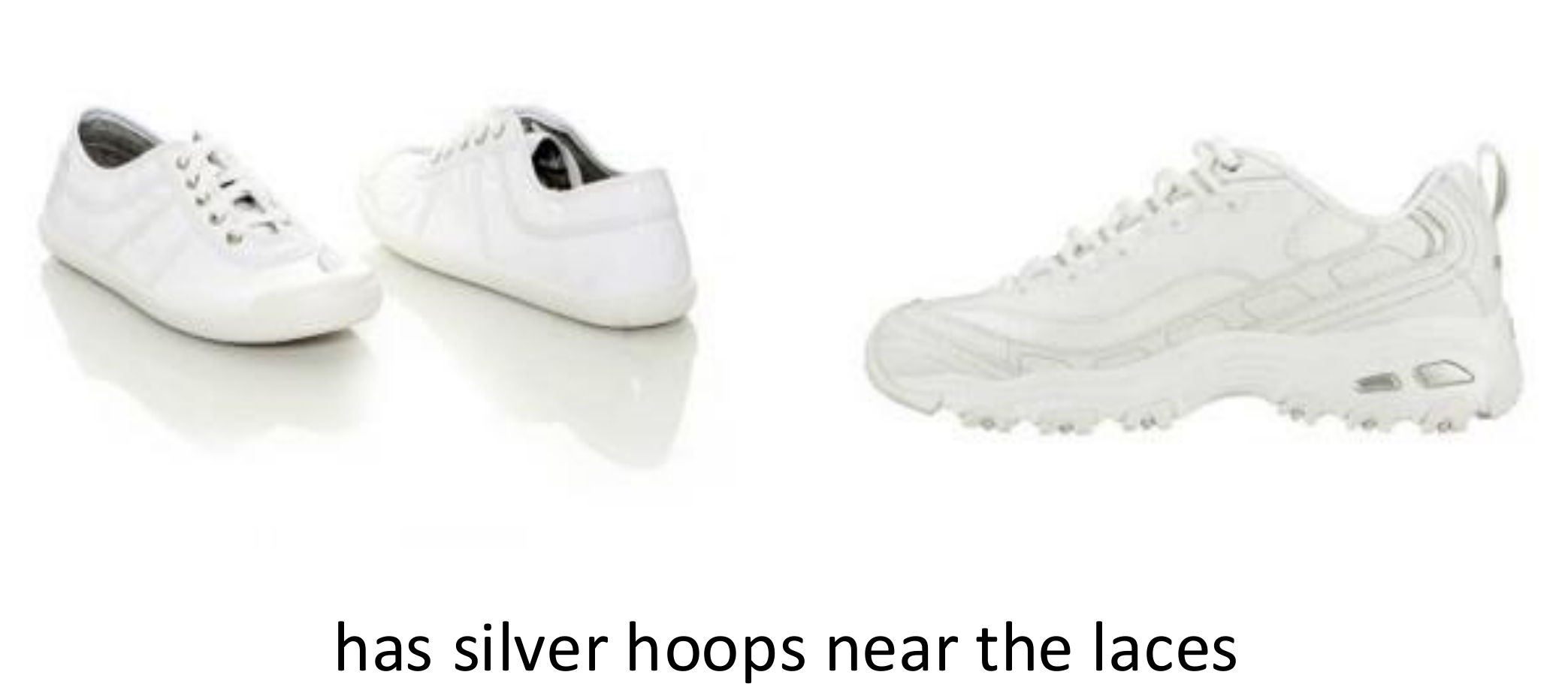}}
\subfigure{
\label{Fig.planes}
\includegraphics[width=0.45\textwidth]{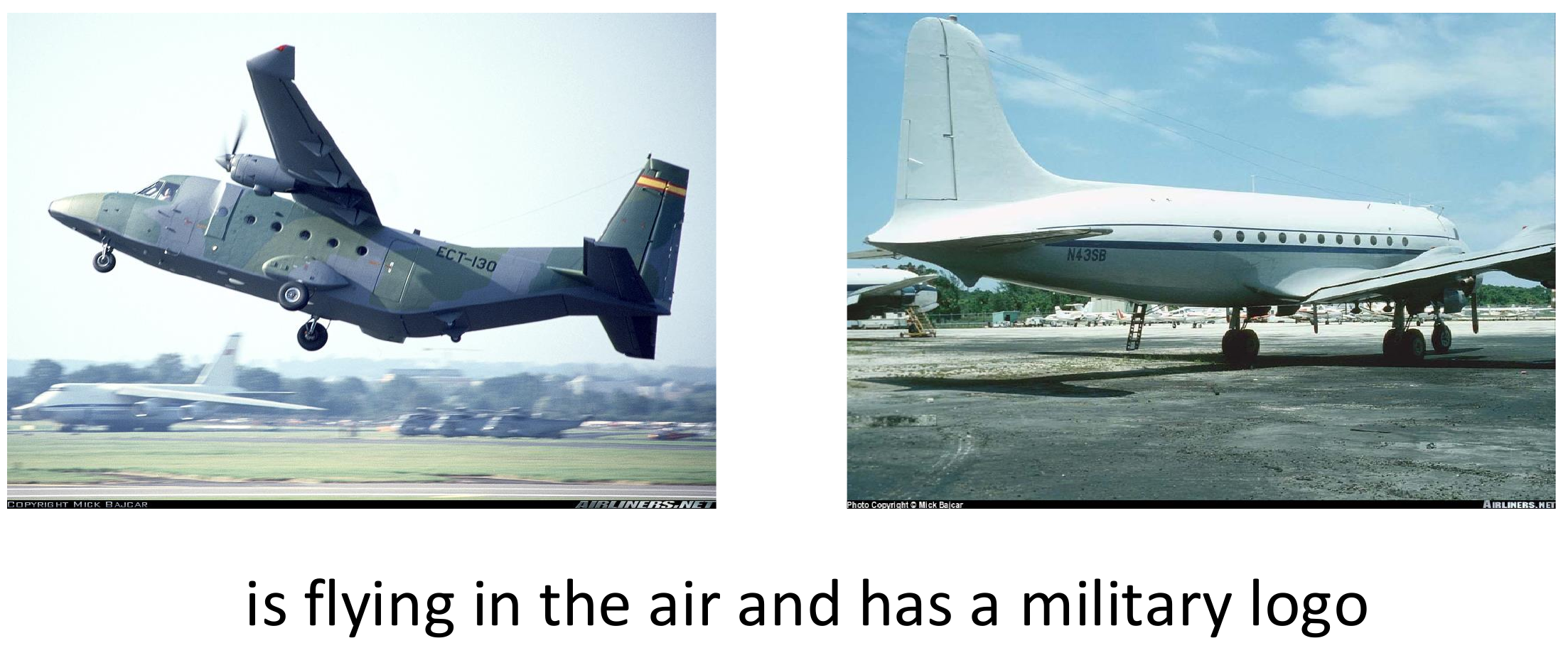}}
\caption{Semantic difference description for image pairs. The left pair is from AMT-20K dataset which is about shoes; the right pair is from OID dataset for planes. Both descriptions only take the difference characteristics shown in the first image into consideration, which is actually how human express.}
\label{Fig.example}
\end{figure}

Difference caption is such a difference telling task between image pair with vast application scenarios: a) guide online shopping with human-computer dialog; b) characterize differences of patient behavior in medical monitoring; c) describe dangerous behavior in traffic. However, relative caption seems to be a rather daunting task at first sight, since we obviously need to conquer  a series of challenges: 1) Absence of dataset for describing pairwise image differences in natural language; 2) A novel framework for processing pairwise visual information; 3) Adaptive evaluation metric (e.g., ``has propeller engine" is also the right caption for Fig~\ref{Fig.planes} but will get a low score in automatic NLP metric).

We researched plenty of popular caption datasets, however they are not capable of covering the need of our task: comprehensive difference in detail; not for entire scene, but for specific object. Datasets in~\cite{hu2016natural,mao2016generation} failed to fit our need obviously: they are proposed to retrieve, not to distinguish. Using caption datasets with individual image is only capable to get "rough" difference caption, which is not what we need. We begin with collecting a dataset of difference caption named AMT 20K dataset by requesting workers to annotate visual difference description $\bm{S}$ with given image pair $(\bm{I_1},\bm{I_2})$ based on the first image $\bm{I_1}$, which is the manner that more like human. Examples are shown in Fig~\ref{Fig.data}. Since that annotations are from non-expert workers, the fine-grained discrepancies is abundant with more subjectivity such as ``a cowboy look'' and ``gold rivet accents''. 

Utilizing the collected AMT 20K dataset, we introduced a series of novel tactics to fuse visual information obtained from two siamese encoders into unified representations, which are prepared for the decoder. Experimental caption results demonstrate the effectiveness of our feature fusing tactics with exquisite discrepancies descriptions generated. The model of our work is designed under the scenario of visual difference caption, contributing to that the input format and evaluation are different with most previous works~\cite{vedantam2017context,hu2016natural,mao2016generation}. Moreover, a referee, which has been proved to be more suitable for our task than automatic NLP metric, is introduced to evaluate our descriptor. We first validate the reliability of our referee, who ranks the correct image with 90.48\% accuracy given ground truth from AMT 20K dataset, while it reached at 82.34\% accuracy given captions generated by our model. 
We also compare our work with the state-of-the art of Su et al.\cite{su2017reasoning} on the OID:Aircraft dataset, achieving a score of 2.88 than their 2.72 evaluated with user study method.

\noindent \textbf{Contributions of this paper.} In summary, our work makes the following contributions:
\begin{itemize}
\item A novel caption problem is to tell the difference in the given image pair and provide human-style description, which can be widely used in various applications. In this paper, we develop a complete system for this problem, containing datasets, caption model and evaluation metric.
\item We introduce a new large-scale dataset for fine-grained discrimination caption named AMT 20K of nearly 32000 image pairs and relative annotation. Besides, an additional dataset named OID:Aircraft is re-annotated with 9400 image pairs based on the OID dataset~\cite{su2017reasoning}. Moreover, a low-cost method for data enrichment is proposed. Fig~\ref{Fig.data} shows some representative samples of both datasets.
\item We propose a series of feature fusing modules for the visual differences caption task, fusing multiple visual representations of either high or low level into one,
that a unified representation is consist of either high or low level features, which can be adopted by most encoder-decoder caption models.
\item In this work, a referee model is introduced to evaluate our relative captioner. Extensive experiments demonstrate that the referee is the most suitable evaluation metric compared to automatic metric and user study.
\end{itemize} 

\section{Related Work}

\textbf{Caption Generation Task.} \label{Caption Generation Task}
Generating descriptions on target image has been extensively studied in both Computer Vision and Natural Language Processing~\cite{rennie2016self,ren2017deep,gan2017stylenet,kiros2015skip}. 
Early researches take advantage of sentence template and heavily hand-designed systems \cite{kulkarni2013babytalk,mitchell2012midge,yang2011corpus}, which limits the application so far as to cause sensitivity to disturbance.
Traditional approaches of captioning image comprise Encoder-Decoder structure and end-to-end training method. 
The complete framework extracts feature representation from pre-trained deep convolutional networks ,then the language model decodes image features to generate descriptive sentences.
LSTM~\cite{hochreiter1997long} and GRU~\cite{cho2014learning} are favored in demonstrating sequential relations among words.
Proposed by Vinyals et al.~\cite{vinyals2015show}, this well-designed structure has been hitherto adopted by many state-of-the-art caption generation frameworks \cite{yao2016boosting,krishna2017dense,liu2016improved}.

Recently, attention-based models~\cite{xu2015show,you2016image,fu2017look,Yang_2016_CVPR,karpathy2015deep,xu2017jointly} have achieved higher performance by a large margin. 
Attentive mechanisms that embedded in neural networks are capable of capturing local regions out of the full image. 
Looking into regional details facilitates recognition to the image, which is in accordance with human intuition~\cite{desimone1995neural}. 
Xu et al.~\cite{xu2015show} proposed two variants of attention models. 
Moreover, they validate the use of attention with state-of-the-art performance.
You et al.~\cite{you2016image} applied a policy that extracts richer information from the image and couples them with an RNN selectively attend on semantic attributes. 
Karpathy et al.~\cite{karpathy2015deep} provided an image-sentence ranking mechanism for visual-semantic alignment. Different from most of the one-image-based caption models, our work contributes by learning the representation of differences between image pairs.

Inspired by human attention as well, V. Mnih et al.~\cite{mnih2014recurrent} introduced a Recurrent Attention Model that a deep recurrent neural architecture iterates image features from previous steps, while DT-RAM~\cite{li2017dynamic} utilized learnable parameters to govern whether stop the computation and output the results at each time step. 
As to our formulation, attention mechanism effects both in descriptions generating and differences emphasizing.

\noindent \textbf{Image difference Detection.}
Though less popularity like relation detection enjoys, present works along with remarkable outcome preview great potential in visual comparison application.
 Yu et al.~\cite{yu2014fine} proposed a local learning-to-rank approach by which the model predicts which one exhibits the attribute more from a fine-grained image pair. 
The team further their work~\cite{yu2015just,yu2016semantic} towards deciding whether these differences in attributes is perceptible, or 
the relative strength of an attribute.
The Attribute-based visual difference has been studied in ~\cite{su2017reasoning} as well.
The determinant of relationships among objects in image understanding is emphasized by Dai et al.~\cite{dai2017detecting} and Cheng et al.~\cite{cheng2018multi} respectively.

Further image difference detection works include generating descriptions afterward, for example, Liu et al.~\cite{liu2018show} and Su et al.\cite{su2017reasoning}. To be specific, the latter team utilized encoder-decoder framework to generate attribute phrases in accordance with image pairs from OID Dataset\cite{Vedaldi2014Understanding}, which we have modified so that apply to our model (Section\ref{sec:Dataset}). 
Such work 

Similarly, Andreas et al.\cite{andreas2016reasoning} and Vedantam et al.\cite{vedantam2017context} treated the second image as a distractor, while the former focused on scene description and the latter features using training data that only describe a concept or an image in isolation.
Different from present works, we emphasise on the flexible syntactical structure of our difference captioner that facilitate generating ``realistic" human-like difference description out of raw image pairs.

\section{Model}

\textbf {Overview.}
Deep Neural Networks are powerful machine learning models that achieve outstanding performance in sequence to sequence translation~\cite{sutskever2014sequence}, image caption~\cite{vinyals2015show,xu2015show} and many other natural language processing fields. In this work, we introduce an amusing issue about telling the difference of image pair in a human-like style. Given a correct difference description $\bm{S}$ and the source image pair ($\bm{I}_1$, $\bm{I}_2$), this issue can be modeled as finding the description that maximizes the following log probability:
\begin{equation}
\label{Equ:probability}
\log p(\bm{S}|\bm{I}_1, \bm{I}_2) = \sum_{t=0}^T \log p(\bm{S}_t|\bm{I}_1, \bm{I}_2; \bm{S}_0, \bm{S}_1, ..., \bm{S}_{t-1})
\end{equation}
, where $\bm{S}_t$ is the word in the description at location $t$, and $\bm{S}$ has $T$ words in total. In the training process, in order to maximize the log probability of a correct description, we set the training objective as:
\begin{equation}
Objective =  - \sum\limits_{i = 1}^N {\log p\left( {{\bm{S}^i}|\bm{I}_1^i,\bm{I}_2^i} \right)}
\end{equation}
, where $N$ denotes the number of the training set. Once training process is completed, we inference difference description by finding the sentence $\bm{\hat S}$ that maximizing the probability above:
${\bm{\hat S}} = \arg \mathop {\max }\limits_{\bm{S}}  p(\bm{S}|\bm{I}_1,\bm{I}_2)$.

\begin{figure*}[!t]
\centering
\includegraphics[height=8.0cm]{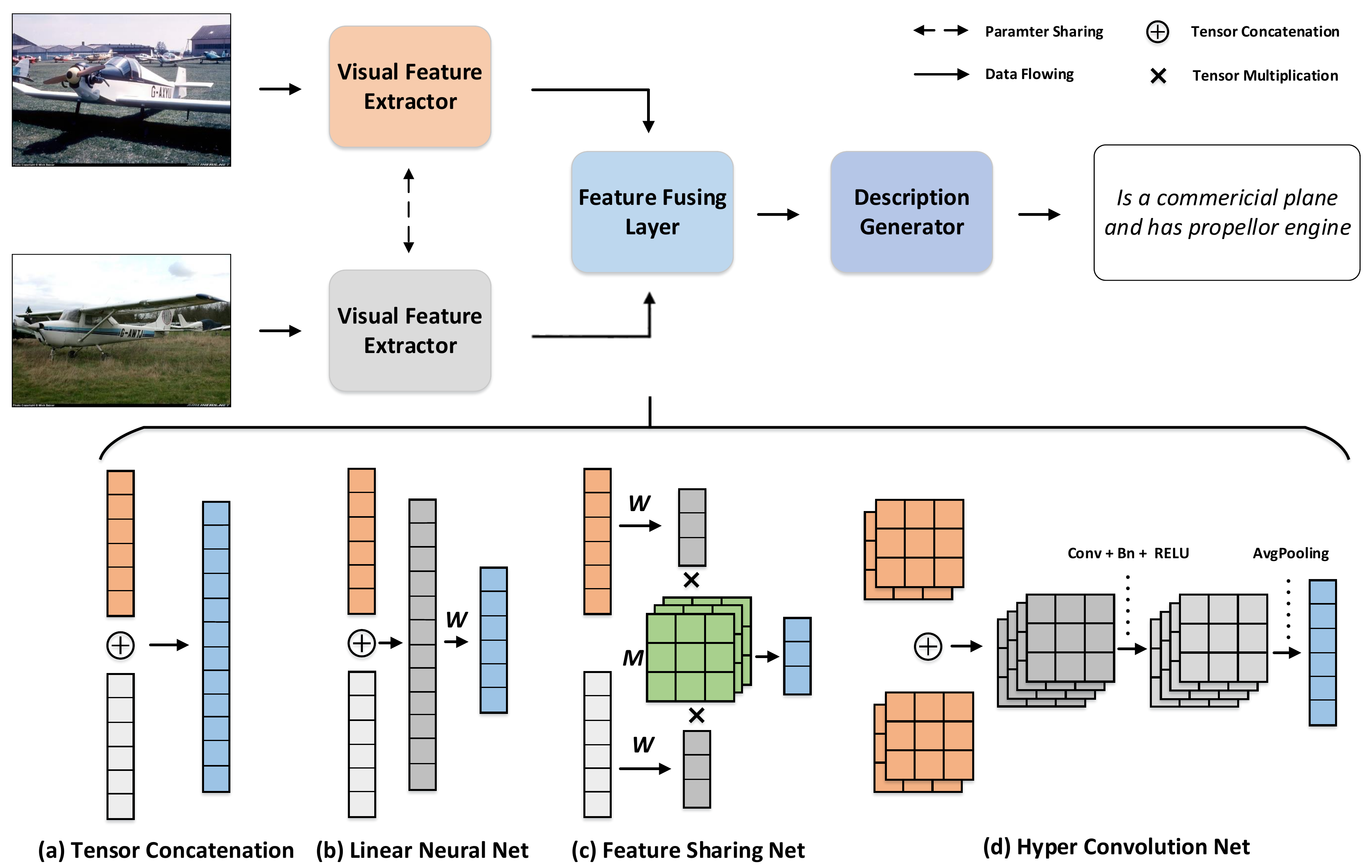}
\caption{Our visual descriptor framework. Pairwise images are fed into Visual Feature Extractors (ResNet\cite{He_2016_CVPR} in our work) respectively which share parameters with each other to obtain visual features. After that, a series of Feature Fusing Tactics are introduced to fuse features from two independent images into one, which is obtained for Description Generator to tell the difference. We illustrated four fusion methods to fusion pairwise features from different point of view, introduced in Section \ref{Fusion}.}
\label{fig:descriptor}
\end{figure*}

In this section, we proposed a pairwise generator model, shown in Fig \ref{fig:descriptor}, which followed the state-of-the-art framework of generating image caption in a features encoder-decoder structure. A feature extractor (Section \ref{Feature Extractor}) is used as encoder to encode images to visual features. These pairwise features are then fused utilizing a series of fusion tactics (described in Section \ref{Fusion}). In the last step, long short-term memory (LSTM) with attention mechanism (Section \ref{Description Generator}) as a decoder is adopted to realize the function shown in Equation \ref{Equ:probability}.

\subsection{Feature Extractor}
\label{Feature Extractor}

As mentioned in Section \ref{Caption Generation Task}, we use a Convolutional Neural Network (CNN) that extracts representations of images. Our particular choice of CNN leads to deep residual learning framework~\cite{He_2016_CVPR}. Known as Deep residual network (ResNet), it features "shortcut connections" that connect lower layers by skipping one or plural layers and added to the output of stacked layers: 
\begin{equation}
\mathbf{y=\mathcal{F}(x,\left\{W_i\right\}) + W_s x}
\end{equation}
, where $\mathcal{F}(x,\left\{W_i\right\})$ is the residual mapping to be learned, and $\mathbf{W_s x}$ is a linear projection that shortcut connections. ResNet, as one of the most competitive image classification model, can provide the decoder with strong and stable representations. Concretely, we juxtaposed two ResNets-101 initialized with a pre-trained model on ImageNet~\cite{Russakovsky2015ImageNet}, which share identical parameters with each other.

Compressed high-level representation benefits category classification task, while properly preserved low-level representation is in favor of fine-grained features like texture and color attributes hence facilitate our model for difference detection. Therefore, both high-level and low-level representations are adopted in our work. 1) We obtain the outputs of the last convolution block as the low-level representation, denoting as $\bm{C}\in \mathbb{R}^{k\times l\times l}$, where $k=2048$ represents the channel number, $l$ represents the width and height of the feature maps; 2) The high-level representation $\bm{fc}\in \mathbb{R}^{k}$ is obtained by operating average pooling on $\bm{C}$. According to these two different represents, we introduce a series of fusion tactics. Now all the visual features are eligible for our fusion model.

\subsection{Fusion tactics}
\label{Fusion}
In this part, the frameworks of four fusion methods will be illustrated, followed by Section \ref{sec:analyze}, where we reveal evaluation results and performance analysis among them. 
The goal of the fusion layer is to combine each pair of feature vectors or matrixs from image pairs into a holistic representation, which can be adopted by mainstream caption models. 
These feature vectors and matrixs are identical with the visual features mentioned in the last paragraph of Section \ref{Feature Extractor}.
As an effort to make a further step, we have experimented on multiple methods that enable us to combine each pair of $\bm{fc}_1/\bm{C}_1$ and $\bm{fc}_2/\bm{C}_2$ into a vector $ \bm{r}$.
Some of them are unconvincing and show up with poor results, while the four fusion methods listed below appear to be promising.

\begin{itemize}
\item \textbf{Tensor Concatenation} Our first move from intuition is to directly concat two feature vectors $\bm{fc}$ into vector $\bm{r}\in \mathbb{R}^{2k}$. Denote $[*]$ as the vector concat or matrix channel-stack manipulation, $\bm r$ is obtained by: 
\begin{equation}
\bm{r} = [\bm{fc}_1,\bm{fc}_2].
\end{equation}

\item \textbf{Linear Neural Net} The second choice of fusion layer is a simple but effective network with affine transformation prior and $\rm ReLU$ activation afterwards. Denoted as: 
\begin{equation}
\bm{r} = {\rm ReLU}(\bm{W}[\bm{fc}_1,\bm{fc}_2]+\bm{b})
\end{equation}
, where $\bm{W}\in \mathbb{R}^{2k\times k}$, and the result $\bm{r}\in \mathbb{R}^{k}$.

\item \textbf{Feature Sharing Net} Thanks to the flexibility of visual feature tensors, it is our third option that $\bm{fc}_1$ and $\bm{fc}_2$ have been affinely transformed before multiplication with feature sharing matrix $\bm{M}$ with a $\rm ReLU$ activation following:
\begin{equation}
\bm{r} = {\rm RELU}\left\{ {\left( {{\bm{W}_1} \cdot \bm{fc}_1 + {\bm{b}_1}} \right) \cdot \bm{M} \cdot \left( {{\bm{W}_2} \cdot \bm{fc}_2 + {\bm{b}_2}} \right)} \right\}
\end{equation}
, where ${\bm{W}_1},\bm{W}_2\in \mathbb{R}^{k\times m}$; parameters of $\bm{M}\in \mathbb{R}^{m\times m\times m}$ will be updated when training. The result $\bm{r}\in \mathbb{R}^{m}$.

\item \textbf{Hyper Convolution} Last fusion method achieves the prime outcome. Visual features with spatial information $\bm{C}$, reserved from last convolution layer of ResNet, is fed into our Hyper Convolution (HC) layer. In HC layer, we concat $\bm{C}_1$ and $\bm{C}_2$ in the nominal channel dimension before a set of $1\times 1$ convolution layers. Moreover, operations that batch normalization and ReLU activation enable boosting of the training procedure. The result $\bm{r}\in \mathbb{R}^{k}$ is denoted as:
\begin{equation}
\bm{r} = {\rm AvgPooling}\left\{ {{\rm RELU}\left( {{\rm Bn}\left( {{\rm Conv}\left( {\left[ {{\bm{C}_1},{\bm{C}_2}} \right]} \right)} \right)} \right)} \right\}.
\end{equation}

\end{itemize}

Once the fused vector $\bm{r}$ is obtained, the description generator is able to decode $\bm{r}$ to tell the difference.

\subsection{Description Generator}
\label{Description Generator}




We leverage a single-layer n-timestep LSTM as our language model as well as RNN decoder, by generating one word at each time step given attention weight($\alpha_t, t = 1,...,n, \alpha_t \in \mathbb{R}^d$ ) and context vector(${{\tilde z}_t, t = 1,...,n, {\tilde z}_{t} \in \mathbb{R}^{v \times d}}$). 

In order to derive ${\tilde z}_t$, two prerequisite variables should be provided, which are $\rho_i$ and $\alpha _i$. Here, $\rho_i, i=1,...,d, \rho_i\in \mathbb{R}^{w}$ are local feature vectors, sliced from $\Upsilon \in \mathbb{R}^{w\times d}$, 
where $\Upsilon$ is computed by average 2d pooling from fused vector $\textbf{r}$ in Section \ref{Fusion}.
Specifically, $d$ is alterable from user setting, deciding how many local regions to scope within the visual feature vector $\bm{r}$. 
Each $\rho_i$ is tied up with a positive weight $\alpha _i$. 
The attention mechanism($f_{att}$), a multilayer perceptron, derives each weight $\alpha _i$ which indicates each annotation vector $\rho_i$'s relative importance, conditioned on previous hidden state (${e_{ti}} = {f_{att}}(\rho_i,{h_{t - 1}})$). Concretely, each $\alpha _r$ is activated by softmax function(${a_{tr}} = {\exp ({e_{tr}})}/({\sum\nolimits_{i = 1}^L {\exp ({e_{ti}})} })$). 

Then, context vector ${\tilde z}_t$ is presented in a conditional expectation while output word probability is acquired given the word embeddings($E{y_{t - 1}} \in \mathbb{R}^{1 \times 1}$, 1 is the size of vocabulary), hidden state($h_t$), and context vector(${{\tilde z}_t}$), as introduced by Bahdanau et al.\cite{bahdanau2014neural} and Xu et al.\cite{xu2015show} respectively. 


In nature, the attentive model is trained on fusion features and ground truth sentences, so as to emphasize the most obvious difference from human intuition.


\section{Discriminating Referee}
\label{Discriminating Referee}

What accessible to other image captioning tasks could be obscure for ours. In addition to applying automatic metrics (Section \ref{Automatic metrics}), we need an alternative method to evaluate our results. The reason for this is that multiple notable differences exist given two images and these objective discrepancies may vary from image to image, but also some key points are indistinguishable for the automatic metric.
For example, the plane image pair in Fig~\ref{Fig.planes} has multiple notable such as that the former plane has less windows and is smaller, which will cause sentences that focus on attributes out of ground truth get a worse point with automatic matrices. Besides, for this image pair, the score of automatic matrices is high if generator inferences "facing right" as "facing left", which is totally wrong in this task.

Also important is that although we introduced methods of caption generator and a referee, it does not necessarily indicate that we are referring to a generative adversarial structure \cite{goodfellow2014generative}.
Using the referee auxiliarily to train our generator models by utilizing it as an adversarial loss shows a visible increase in successful judgments of the referee, whereas, it has raised little progress in automatic metric evaluations and also led to poor quality texts according to human study.
Moreover, allowing the referee to play both roles
in training and evaluation process provides inappropriate advantages to the referee, which seems like cheating.
The ultimate goal of our task is to improve the readability and precision of generated captions, rather than improve the accuracy of our referee's judgments, for example.

\begin{figure*}[!tp]
\centering
\includegraphics[height=5.2cm]{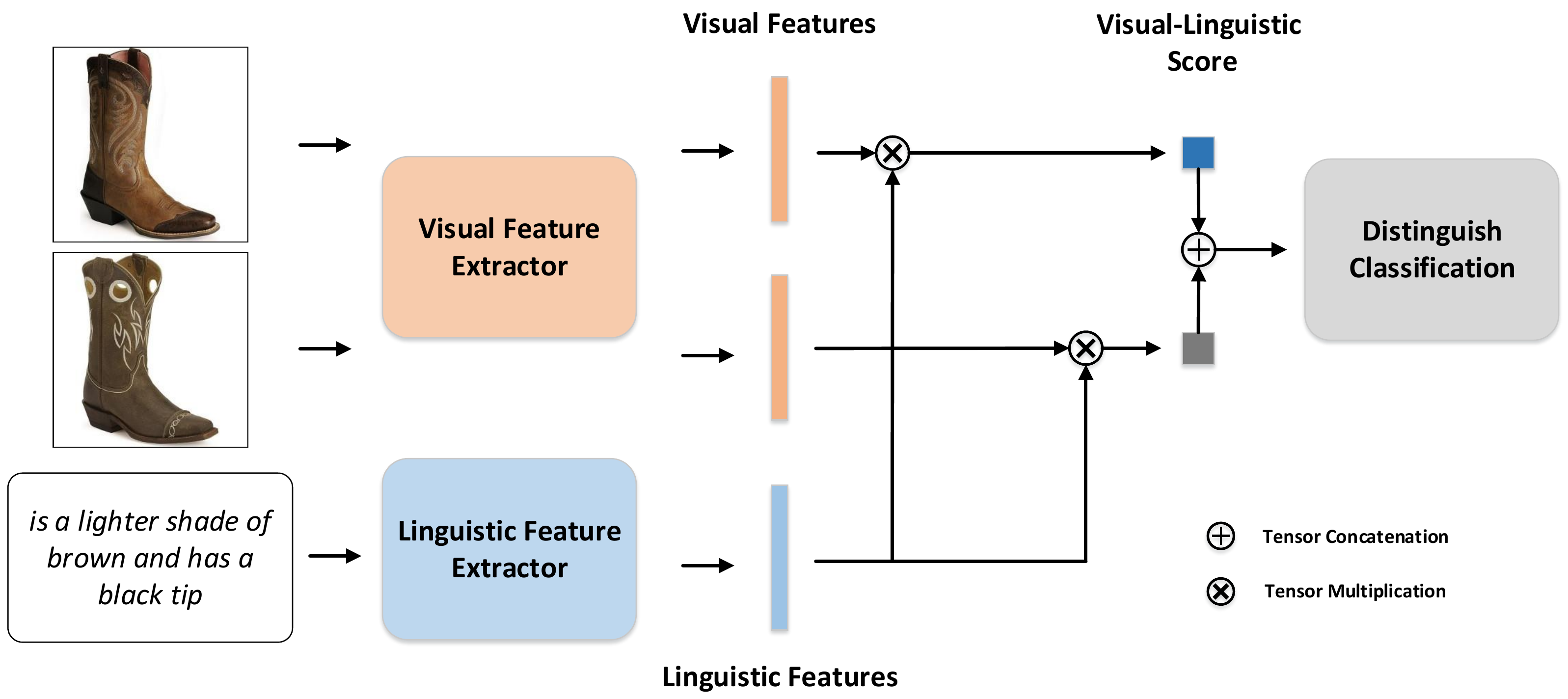}
\caption{Our referee framework. Visual feature pair and linguistic feature are obtained from VGG-16~\cite{Simonyan2014Very} and Text CNN~\cite{Kim2014Convolutional} respectively. Then we define the visual-linguistic score by vector multiplication between visual and linguistic feature. Concatenating of two scores, the final score vector is adopted with a distinguish classification to judge which image is described by our descriptor.}
\label{fig:referee}
\end{figure*}

Motivated by Su et al.~\cite{su2017reasoning}, we introduced a discriminating referee for evaluation process. In the referee task, the training data is the same with the generator task. The referee takes $\bm{I_1}$, $\bm{I_2}$ and $\bm{S}$ as inputs, and inference index $y$ as ground truth. $y=0$ means the difference description $\bm{S}$ is based on $\bm{I_1}$. We randomly exchange the position of $\bm{I_1}$ and $\bm{I_2}$ to make ${y}\sim B(1,0.5)$. The overall referee framework shows in figure~\ref{fig:referee}. Annotate the visual net (ResNet-18~\cite{He_2016_CVPR}) as $\nu$ and the sentence-to-vector embedding net (Text CNN~\cite{Kim2014Convolutional}) as $\omega$, the visual-linguistic score $s_{vl}$ formulates as:
\begin{equation}
s_{vl} = v^T l = \nu  (\bm{I'}; \theta _\nu)^T\omega (\bm{S}; \theta _\omega)
\end{equation}
, where $v$ represents visual feature, $l$ represents linguistic feature. $\bm{I'}$ shows the order of images that appears to referee. Network $\nu$ and $\omega$ have linear layer in the last layer respectively to map vectors of different size into the same size $k=1024$ before vector product. Therefore, we get the probability of $\bm{S}$ based on $\bm{I_1'}$:
\begin{equation}
\begin{split}
p\left( {y = 0|{\bm{I}_1'},{\bm{I}_2'},\bm{S}} \right) = {{\exp \left( {s_{vl}^1} \right)} \mathord{\left/
 {\vphantom {{\exp \left( {s_{vl}^1} \right)} {\left( {\exp \left( {s_{vl}^1} \right) + \exp \left( {s_{vl}^2} \right)} \right)}}} \right.
 \kern-\nulldelimiterspace} {\left( {\exp \left( {s_{vl}^1} \right) + \exp \left( {s_{vl}^2} \right)} \right)}}
\end{split}
\end{equation}
. The target of referee is to find $\Theta  = \left\{ {{\theta _\nu},{\theta _\omega}} \right\} $ that maximum the probability as below:

\begin{equation}
    \begin{split}
    \Theta = \mathop {\arg \max }\limits_\Theta \{ {\left( {1 - y} \right)p\left( {y = 0|{\bm{I}_1'},{\bm{I}_2'},\bm{S}} \right) +} 
    {y\left( {1 - p\left( {y = 0|{\bm{I}_1'},{\bm{I}_2'},\bm{S}} \right)} \right)} \} .
    \end{split}
\end{equation}

It translates reference game into a two-category classification task, making it trainable with the same data as our descriptor. The referee is now able to judge whether the description our model generates tells the difference or not, no matter whether the description has attributes outside the ground truth. Besides, it's sensitive to those key points mistakes.



\section{Experiments}

We evaluate our generator and referee on two different datasets - one is collected by us and the other is collected based on existing dataset. In the Section~\ref{sec:Dataset}, the AMT 20K dataset, one major contribution of our paper, will be introduced. Section~\ref{sec:metrix} introduce the exploration of the best evaluation metric in the difference caption task. Model study are shown in Section~\ref{sec:setting}. Section~\ref{sec:analyze} demonstrate the experiment results and compare with the state of the art respectively. 

\subsection{Dataset details}
\label{sec:Dataset}




There are comparatively few datasets for fine-grained difference attribute phrases in computer vision,


the situation is even worse for who focuses on difference caption of the image pair. As far as we know, there is no such fine-grained difference description dataset for our task at present. As a pathfinder in image difference caption, we introduce two datasets, one is collected by ourselves, and the other is transformed from the OID aircraft attribute phrases dataset~\cite{su2017reasoning}.

\begin{figure}[!tp]
\centering
\subfigure{
\label{Fig.annotation}
\includegraphics[width=0.3\textwidth]{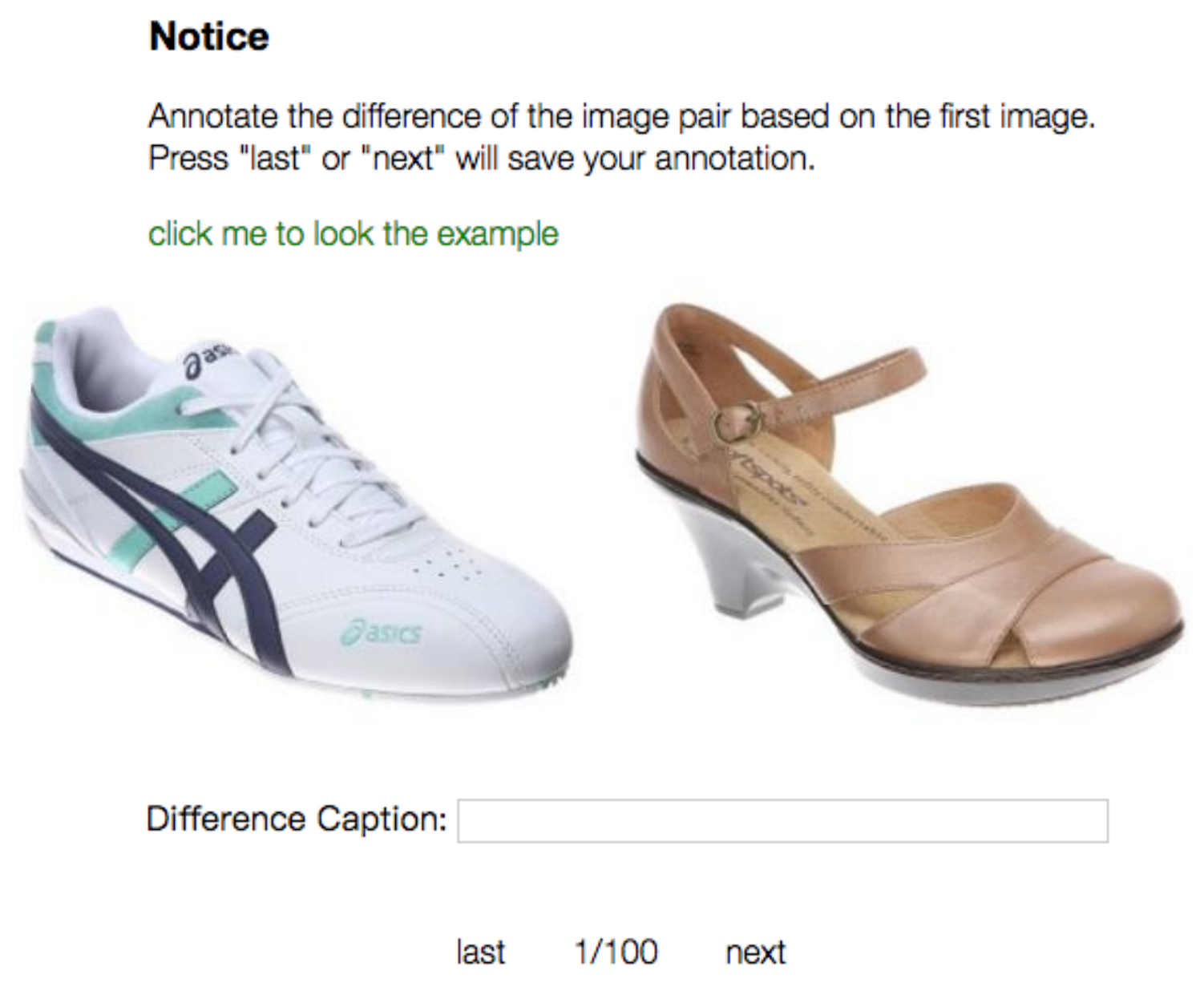}}
\subfigure{
\label{Fig.amt_20k}
\includegraphics[width=0.3\textwidth]{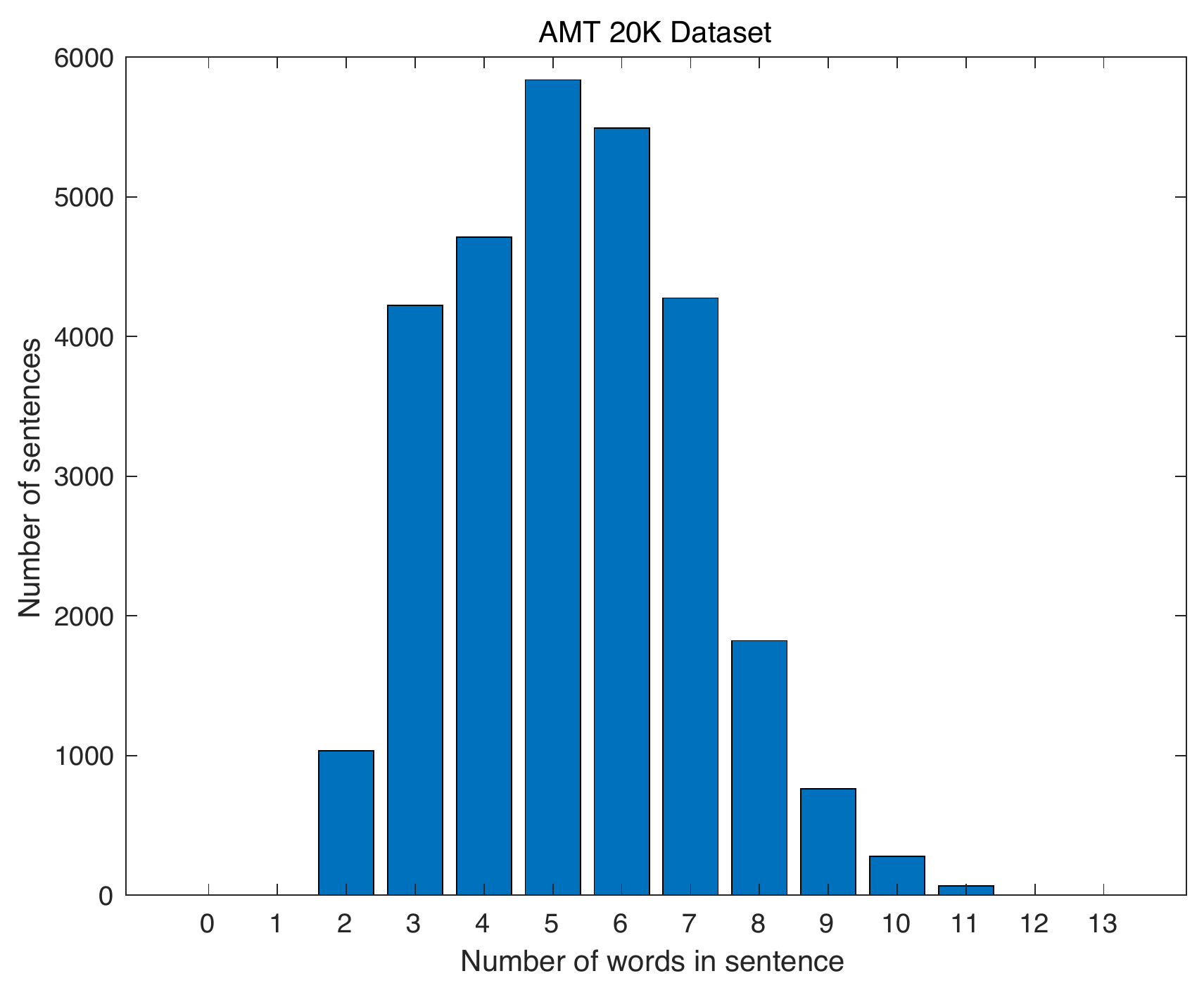}}
\subfigure{
\label{Fig.oid}
\includegraphics[width=0.3\textwidth]{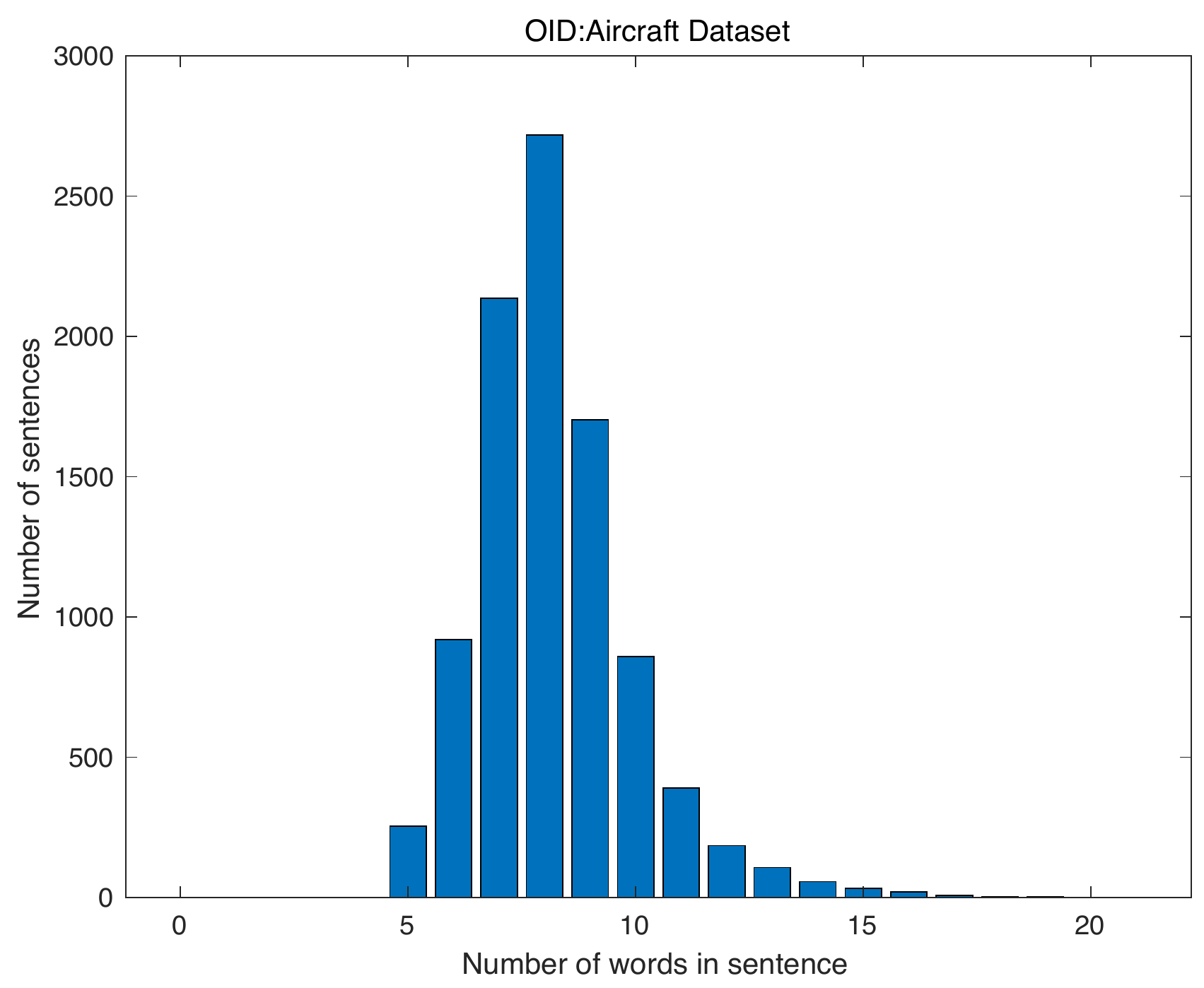}}
\caption{(a) The left figure is the annotation interface used to collect difference captions; (b) The middle figure is the sentence length histograms of images in AMT 20K dataset; (b) The right figure is the sentence length histograms of images in OID:Aircraft dataset.}
\label{Fig.dataset_distribution}
\end{figure}

\textbf{AMT 20K Dataset.} 
UT Zappos50K dataset~\cite{finegrained} is a large shoe dataset for pairwise attributes comparison task, consisting of 50,025 catalog images collected from Zappos.com. Based on those 14764 chosen images from UT Zappos50K dataset, we annotate image pairs with different descriptions to collect our own AMT 20K dataset. The annotation method is shown in Fig~\ref{Fig.annotation}, which is released to a group of workers to annotate on the Internet. The length of collected annotation is in the range between 1 to 13, with an average number of 5.28, shown in Fig~\ref{Fig.amt_20k}.

The annotations are organized into 26710 image pairs (12416 images) in training set, and 5041 pairs (2348 images) in test set. Since that image pairs are annotated by different workers without any constraint in advance such as attribute range, sentence pattern and so on, the difference descriptions are diverse of open-ended "realistic" human-style. Besides, the annotations contain various attributes/key points, such as color, texture, accessories, size and so on, which enriches the visual comparison details compared to those attribute-based datasets. More statistics of AMT 20K dataset and its comparison with other datasets are given in Table~\ref{table:length}.

\setlength{\tabcolsep}{4pt}
\begin{table*}
\begin{center}
\caption{Comparison of our AMT 20K dataset with existing datasets}
\label{table:length}
\begin{threeparttable}
\begin{tabular}{|c|c|c|c|c|c|c|c|}
\hline
Dataset & \tabincell{c}{Resolution\\ (w$\times$h)} & ImgNum & PairNum & Max\tnote{1}  & Min\tnote{1}  & Avg\tnote{1}  & SPP\tnote{1}  \\
\hline
\hline
AMT 20K  & 280$\times$280 & 14764 & 31751 & 13 & 1 & 5.28 & 1\\
\hline
OID:Aircraft & 1200$\times$756\tnote{2}  & 7541 & 9400 & 21 & 5 & 8.22 & 1\tnote{3} \\
\hline
\end{tabular}
\begin{tablenotes}
\footnotesize
\item[1] Max is the maximum sentence length; Min is the minimal sentence length; Avg is the average sentence length; SPP: sequence number per pair.
\item[2] The Resolution has a random error of 0-20 pixels.
\item[3] The OID dataset refer to the vision of difference captions.
\end{tablenotes}
\end{threeparttable}
\end{center}
\end{table*}
\setlength{\tabcolsep}{1.4pt}

\textbf{OID:Aircraft Dataset.} Aiming to investigate models that understand fine-grained object categories with rich descriptions, Vedaldi et al.~\cite{Vedaldi2014Understanding} introduced OID Aircraft Dataset of 7,413 airplanes annotated with attributes according to components. Additionally, Su et al.~\cite{su2017reasoning} derived annotations on images from OID Aircraft dataset, picked images pairs uniformly at random within the OID aircraft dataset, and relied on human annotators to discover the space of descriptive attribute phrases. The annotations are organized into 4700 image pairs (1851 images) in training set, 2350 pairs (1730 images) in validation set, and 2350 pairs (2705 images) in test set. 

In our task, since the attribute phrases are similar to sentences, we utilize a series of templates to map two or three attribute phrases to difference captions, according to the word formation and category of phrases. Furthermore, obeying certain probability distribution,  generated sentences alter both in length and structure for the sake of descriptive variety. Once the transformation is completed, three workers refine the sentences generated by templates. By this tactic, difference captions are obtained in a convenient way. The split of this dataset is followed the previous work.

\textbf{Data enrichment.}
\label{Data enrichment} 
As mentioned above, the main challenges of difference caption datasets are the costliness for data collection, which limited the scale. One contribution of our paper is the proposal of a low-cost method for data enrichment.
The simplest way is template-based caption by generating template sentence with attribute words or phrases. However, it contradicts our goal that promote natural language generation of artificial intelligence. Instead, we employ a neural language style transfer methods.

Style transfer has been popular in image processing \cite{gatys2015neural,ulyanov2016texture,johnson2016perceptual}. Shen et al. \cite{shen2017style}, taking a further step on textual style transfer
, formulated a shared latent content space($\mathcal{Z}$) between two sentence domains($\mathcal{X}_1, \mathcal{X}_2$), and sentences of contrary sentiment that respectively compose $\mathcal{X}_1$ and $\mathcal{X}_2$ could be transferred bi-directionally.
Thus, we utilized a lingual model for sentence completion.




\setlength{\tabcolsep}{4pt}
\begin{table}
\footnotesize
\begin{center}
\caption{Comparison between template-based captions and style transferred captions}
\label{table:style}
\begin{tabular}{|c|c|}
\hline
Template-based & Neural style transfer\\
\hline
\hline
has white color and is a big plane & is a bigger plane with white color\\
\hline
has a propeller engine and has multiple windows & has more windows with a propeller engine \\
\hline
\end{tabular}
\end{center}
\end{table}
\setlength{\tabcolsep}{1.4pt}

First, we use attributes from original UT Zappos50K and OID dataset to construct sentences of template-based structure (See left column of Table \ref{table:style}) as $\mathcal{X}_1$, while difference descriptions have been provided by AMT 20K and updated OID dataset as $\mathcal{X}_2$.

Then, it is trained on two sentence domains following classical training procedure.
Finally, obtaining translations from raw phrases into coherent and fluent descriptions (See right column of Table \ref{table:style}) by sending a larger set of template-based sentences to the best-trained model. Despite that rare grammatical errors do occur, it does not necessarily preclude the fruitful achievement that marks effectiveness of our model.


\subsection{Evaluation Metrics}
\label{sec:metrix}

To quantitatively evaluate the generated sentences, we introduced three evaluation metrics for different objectives in our work, namely automatic metric for machine translation, a discriminating referee and user study method. 


\textbf{Automatic metric.} 
\label{Automatic metrics}
Like other common caption generation tasks, we deploy automatic metric for results reporting. Specifically, we tested on BLEU\cite{papineni2002bleu} and ROUGE\cite{lin2004rouge}. They are essentially based on computing unigram or n-gram from the candidate sentence that are found in the reference sentence.
Additionally, BLEU multiplied by a brevity penalty to prevent very short candidates from receiving too high a score while METEOR using the harmonic mean to combine precision and recall, the latter which is proportional to the ratio that the number of unigrams in the candidate to the number of ` in the reference.


\textbf{Discriminating referee.} As introduced in Section \ref{Discriminating Referee}, an alternative evaluation method is to use a carefully designed model, where generated sentences $\bm S$ are given information and how well the referee could distinguish within the image pair $({\bm I_1'},{\bm I_2'})$ whose position is exchanged randomly. Evolved from automatic metric, our referee focuses on whether the vital information is attended by the generator. For example, candidate sentence `the plane on the left is a commercial plane' resemble reference sentence `the plane on the left is a military plane' and scores high in the automatic metric such as BLEU. Unfortunately, it utterly misapprehends the use of the plane in the current image.

GT results in Table.~\ref{table:results} and Table.~\ref{table:comparison} show the accuracy of our referee in the task of $({\bm I_1'},{\bm I_2'},{\bm S})\to \left\{ {0,1} \right\}$, compared with human referee. Due to the noise of datasets and the error of human judgment, the accuracy of human referees is slightly lower than $100\%$. The accuracy of our discriminating referee reaches nearly $90\%$, which is a qualified result to the duties of the referee. The accuracy of the referee in AMT 20K dataset are higher than that in the OID:Aircraft, thanks to the more abundant visual difference in AMT 20K dataset.

\begin{figure*}[!tp]
\centering
\includegraphics[height=2.6cm]{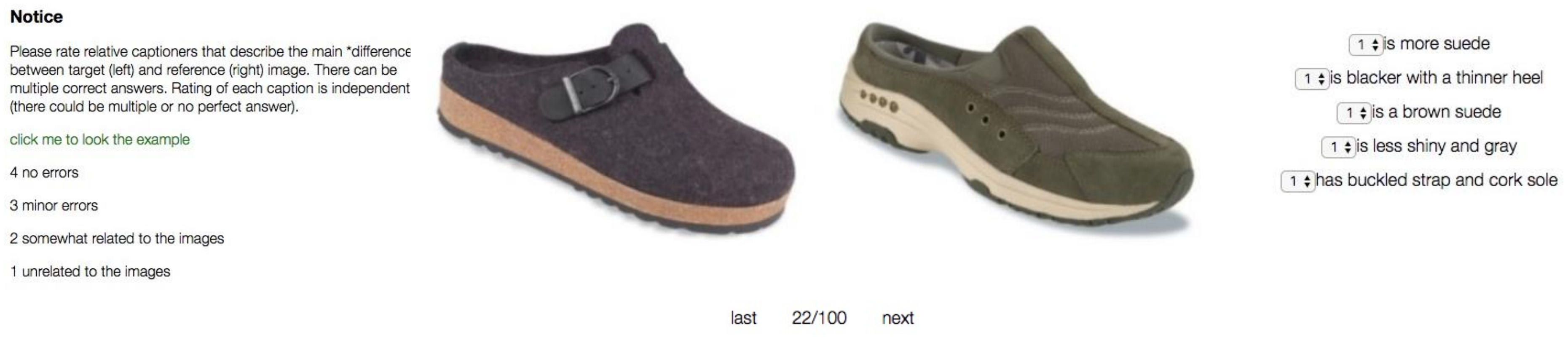}
\caption{User study interface. According to the image pair, workers are asked to rank difference captions in the left of figure with a range of 1 to 4, where 4 is the highest rank for worker's intuition.}
\label{fig:user_study}
\end{figure*}

\textbf{User study.} We resort to human evaluation as well, which is extensive and the most reliable though time-consuming. We leverage a well-designed web page in Fig~\ref{fig:user_study} and request the workers to evaluate each generated sentence of the image pair. Each sentence rank is independent, indicating that there could be multiple or no perfect answer. We took the arithmetic mean of scores from a group of graders. In this work, ten workers are involved for the evaluation of both datasets with 100 data pairs per worker.

\subsection{Experiments Settings}
\label{sec:setting}

In this work, we run our experiments on five GEFORCE GTX 1080 Ti GPUs, and the implementation framework is Pytorch~\cite{paszke2017automatic}. More details are described in the following.

\textbf{Generator Model.} In the feature extractor, the output of ResNet-101 $C \in \mathbb{R}^{2048\times l \times l}$ and ${\bm fc}\in \mathbb{R}^{2048}$, where $l=14$. In the Feature Sharing Net, $m=128$ for sharing matrix $M$. As to the one-layer LSTM, the hidden size is set to 512, and the encoding size of vocabulary is 512. When the attention module is used, the hidden size of the attention MLP is also 512.

During the training process, the ResNet-101 is initialized with weights pre-trained on ImageNet dataset~\cite{Russakovsky2015ImageNet}. Using Adam optimizer~\cite{kingma2014adam} with initial learning rate $lr=0.0004$, $\beta_1=0.9$, $\beta_2=0.999$. The training process is persisted 100 epoches with 64 batch size. During the testing process, we use beam search with beam size 10 to generate difference captions.

\textbf{Discriminating Referee.} Both visual feature and linguistic are mapped to $1024$ vector. In text CNN, there are three kernels with size 3, 4, 5 respectively, with 100 kernel number in common. During the training process, the ResNet-18 is also initialized with pre-trained weights. We use Adam optimizer with initial learning rate $lr=0.001$. Batch size is set to $16$ and max epoch number is $50$. There is a batch normalization layer~\cite{ioffe2015batch} in the last of Text Net.

\textbf{Data augmentation.}
\label{Data augmentation}
Difference caption data are in the form of coupled images along with the corresponding text description, making the data acquisition difficult compared to other tasks. In order to conquer the over-fitting issue, we apply data augmentation on both datasets. For both datasets, each image is randomly cropped 10 pixels in the width and height respectively. In the AMT 20K dataset, random horizontal flips are employed additionally thanks to the lack of `left' and `right' attributes. Besides, we add the identical image pair data (3000 pairs for AMT dataset and 1000 for OID dataset) to improve the robustness of our model. The identical image pair data don't be augmented. Besides, both datasets are enriched by 10\% utilizing the method mentioned before.

\setlength{\tabcolsep}{4pt}
\begin{table*}
\begin{center}
\caption{Evaluation on different caption models with AMT 20K dataset}
\label{table:results}
\begin{threeparttable}
\begin{tabular}{|c|c|c|c|c|c|c|c|}
\hline
Models & BLEU-1 & BLEU-2 & BLEU-3 & BLEU-4 & ROUGE\_L & Referee & \tabincell{c}{User\\ study}\\
\hline
\hline
GT   & - & - & - & - & - & 89.15\% & 3.64\\
\hline
\hline
SATTC   & 0.17 & 0.06 & 0.02 & 0.01 & 0.20 & 71.49\%  & 2.43\\
\hline
SATLNN   & 0.18 & 0.06 & 0.02 & 0.01 & 0.21 & 76.45\% & 2.68 \\
\hline
SATFSN   & 0.18 & 0.06 & 0.02 & 0.01 & 0.22 & 76.16\% & 2.64 \\
\hline
SATHC   & 0.21 & 0.08 & 0.03 & 0.01 & 0.24 & \textbf{81.77\%} & \textbf{2.83} \\
\hline
\end{tabular}
\begin{tablenotes}
\footnotesize
\item[1] GT: ground truth; SAT~\cite{xu2015show}: show, attend and tell model; SATTC: Tensor Concatenation Net with SAT; SATLNN: Linear Neural Net with SAT; SATFSN: Feature Sharing Net with SAT; SATHC: Hyper Convolution Net with SAT.
\end{tablenotes}
\end{threeparttable}
\end{center}
\end{table*}
\setlength{\tabcolsep}{1.4pt}

\begin{figure*}[!tp]
\centering
\includegraphics[height=5.4cm]{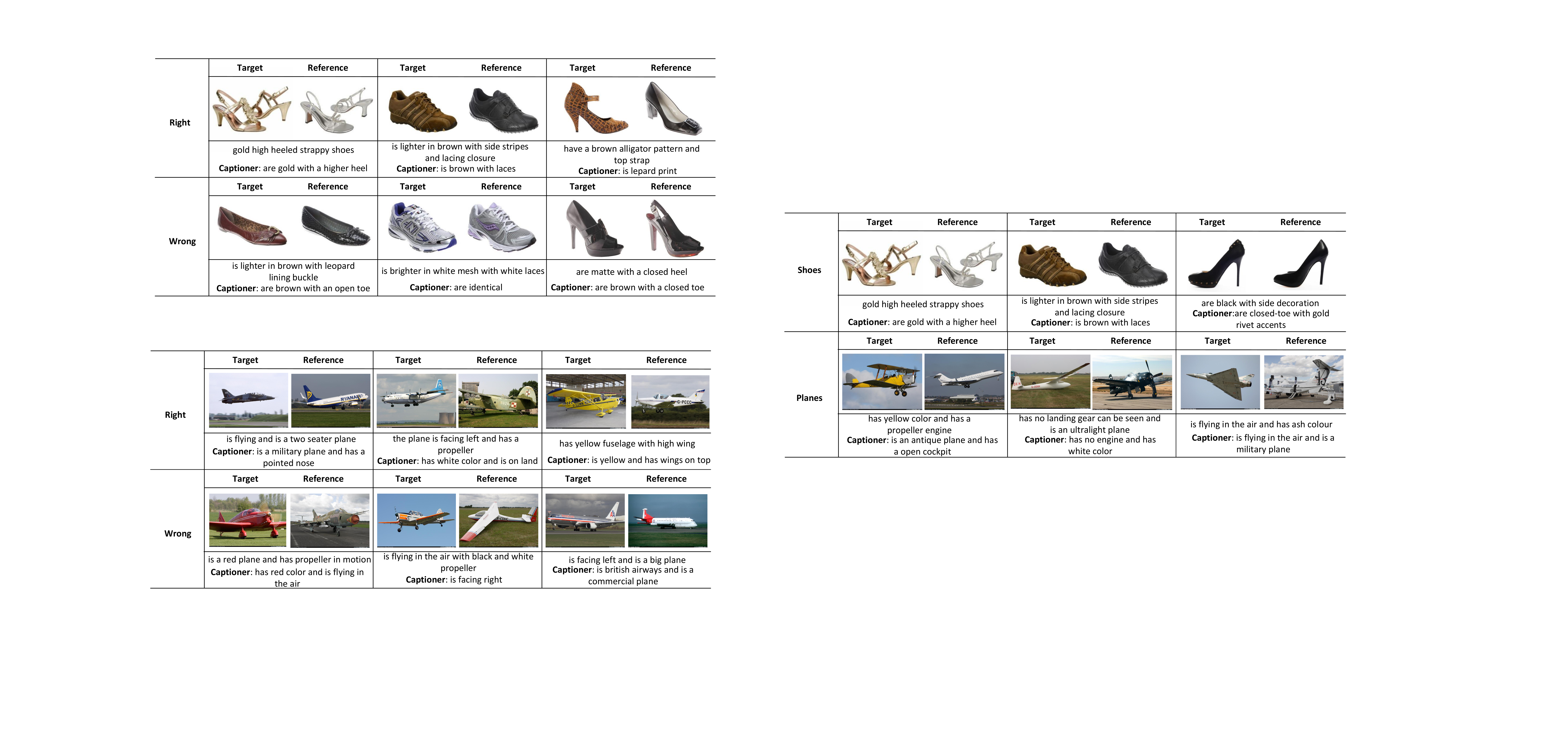}
\caption{Difference caption results by our SATHC model. In each pair, the upper sentence is the discrepancy description collected from human annotators, which we regarded as ground truth, while the lower sentence is the generated caption. }
\label{fig:results}
\end{figure*}

\subsection{Model Analyze}
\label{sec:analyze}

\noindent \textbf{Results on the AMT 20K dataset}

Table~\ref{table:results} shows the results of different methods on the AMT 20K dataset. We utilize various evaluation metric to evaluate our captioner, but take the user study rank as the benchmark. The value of BLEU and ROUGE\_L metric are low in all models, and sightly reflect the performance compared to the referee and user study. The reason is that the discrimination descriptions are various for one pair image. Besides, the captions are usually small due to the minor differences. Therefore, we use referee and user study rank instead automatic metric to evaluate captioner. The Hyper Convolution Net achieves the best performance in both Referee metric and human rank, reaching 82.34\% and 2.88 respectively compared to the SATTC model of only 72.54\% and 2.48 respectively. It's turned that representation fusion in the low-level is better than the high-level.

\noindent \textbf{Comparison with the state of the art on the OID dataset}
\label{sec:comparison}

We compare our results with the most recent state-of-the-art image difference describing task of Su et al.~\cite{su2017reasoning}, by user study method. They trained a \emph{simple speaker}(SS) that takes a single image as input and produces a description, and a \emph{discerning speaker} (DS) that takes two images as input and produces a single (or a pair of) description(s). Given that they only generated short phrases, automatic metric in Section \ref{Automatic metrics} is naturally not applicable. Although we both developed a player-referee demo, listeners of Su et al. and Discriminating Referee of our group are trained on different datasets with discrepancy, and hence evaluations made by computer referee is not comparable. Finally, comparison of accuracy marked by human graders, is shown is Table~\ref{table:comparison}, illustrating that our SATHC model achieved 2.88, compared to 2.72 of DS~\cite{su2017reasoning}. Table~\ref{table:bleu_oid} demonstrates the traditional evaluation matrix result of the SATHC method, which is higher than it in AMT dataset due to the sequence per image is five in OID dataset, but only one in AMT dataset.

\setlength{\tabcolsep}{4pt}
\begin{table*}
\begin{center}
\caption{Comparison of user study results with other methods}
\label{table:comparison}
\begin{tabular}{|c||c||c|c|c|c|c|c|}
\hline
Dataset & GT & SS & DS & SATTC & SATLNN & SATFSN & SATHC\\
\hline
\hline
Referee   & 90.48\% & - & - & 72.54\% & 77.82\% & 77.66\% & \textbf{82.34\%} \\
\hline
User study   & 3.71 & 2.65 & 2.72 & 2.48 & 2.65 & 2.68 & \textbf{2.88} \\
\hline
\end{tabular}
\end{center}
\end{table*}
\setlength{\tabcolsep}{1.4pt}

\setlength{\tabcolsep}{4pt}
\begin{table*}
\begin{center}
\caption{Traditional evaluation matrix of the SATHC method}
\label{table:bleu_oid}
\begin{tabular}{|c|c|c|c|c|c|}
\hline
Matrix & ROUGE\_L & BLEU-1 & BLEU-2 & BLEU-3 & BLEU-4\\
\hline
\hline
SATHC   & 0.54 & 0.70 & 0.55 & 0.41 & 0.29\\
\hline
\end{tabular}
\end{center}
\end{table*}
\setlength{\tabcolsep}{1.4pt}

\noindent \textbf{Generation Results}

\begin{figure*}[!tp]
\centering
\includegraphics[height=2.7cm]{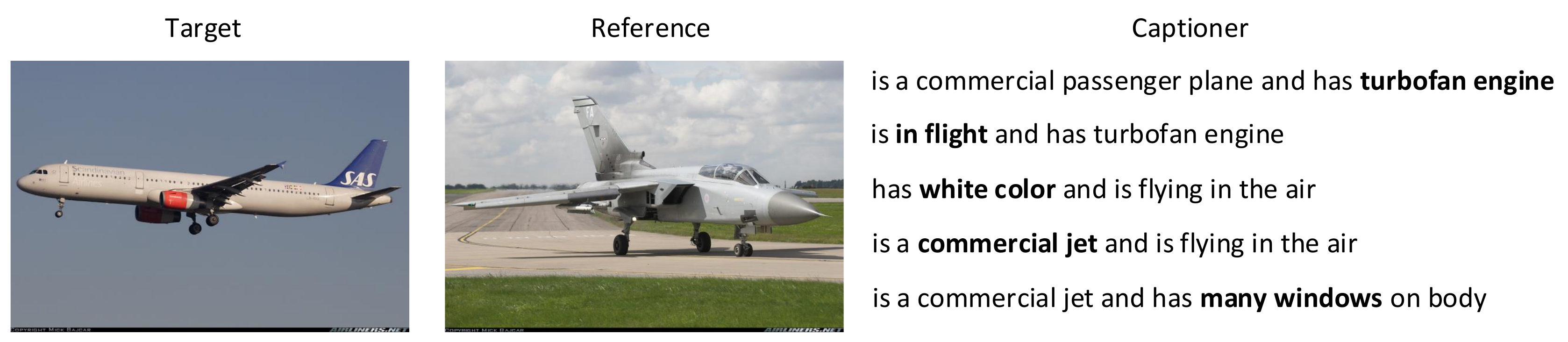}
\caption{Five-best sentences beam searched by our SATHC model with beam size setting to 10. Bold phases indicate a novel discrepancy different from other generations.}
\label{fig:diversity}
\end{figure*}

In each pair in the figure above, the upper sentence is the discrepancy description collected from human annotators, which we regarded as ground truth?, while the lower sentence is the generated caption
The saliency of our model is demonstrated as: a) It is as careful as human annotators that tiny rivet could be named along with multiple other attributes (3rd of Shoe pairs in Fig.~\ref{fig:results}).  b) In addition to evident differences such as color, it is capable of discovering subtle but instructive differences annotations even more distinct than the ones from datasets (1st of Plane pairs in Fig.~\ref{fig:results}). This figure also illustrates that n-gram recall-oriented automatic metric is inappropriate for our task, because the model is permitted to disagree on differences labeled by annotators and thus receiving poor accuracy, even if it is actually correct. Furthermore, in Fig.~\ref{fig:diversity}, diverse relative captions from 5-best beam searching seek out the discrepancy independently, with various visual points. It's proved that our SATHC model has learned something deeper with visual-linguistic representation, and is capable of providing a healthy diversity.

\section{Conclusion}
We have introduced AMT 20K, a large dataset of differential descriptions on fine-grained image pairs. Based on its data, we proposed an encoder-decoder framework with various feature fusing tactics for image difference detection and description generation. Moreover, we enriched our dataset with textual style transfer method unprecedentedly, proved to be reliable as the substitution of human annotation.
Overall, our model reached outstanding results on difference caption task and achieved state-of-the-art performance. In the future work, we would like to introduce more competent relative caption models such as Generative Discriminative structure. Besides, more data will be collected for our AMT Dataset.

\section*{References}

\bibliography{mybibfile}

\end{document}